\documentclass[letterpaper, 10 pt, conference]{ieeeconf}  

\IEEEoverridecommandlockouts                              

\usepackage{hyperref}
\hypersetup{
    colorlinks=true,
    linkcolor=black,    
    urlcolor=blue,
    }
\usepackage{algorithm}
\usepackage{algpseudocode}
\usepackage{graphicx}
\usepackage{subcaption}
\usepackage{booktabs}
\usepackage{multirow}
\usepackage{epsfig} 
\usepackage{amsmath} 
\usepackage{amssymb}  
\usepackage{mathrsfs}
\usepackage{cleveref}
\usepackage{color}
\usepackage{calc}
\usepackage{dsfont}
\usepackage{bm}
\usepackage{float}
\usepackage{url}
\usepackage{xspace}
\usepackage[dvipsnames]{xcolor}

\newcommand{\figref}[1]{Fig.~\ref{#1}}

\usepackage{amsfonts} 
\title{\LARGE \bf Post-Grasp Kinematic Repair for Robotic Insertion via Object-in-Gripper Reorientation}

\author{Haegu Lee and Christoffer Sloth
\thanks{ All authors are with The Maersk Mc-Kinney Moller Institute, University of Southern Denmark, Denmark {\tt\small \{haeg, chsl\}@mmmi.sdu.dk} \tt\small}}

\IEEEaftertitletext{%
    \begin{minipage}{\textwidth}
        \centering
        \vspace{-0.5em}
        \includegraphics[width=0.95\linewidth]{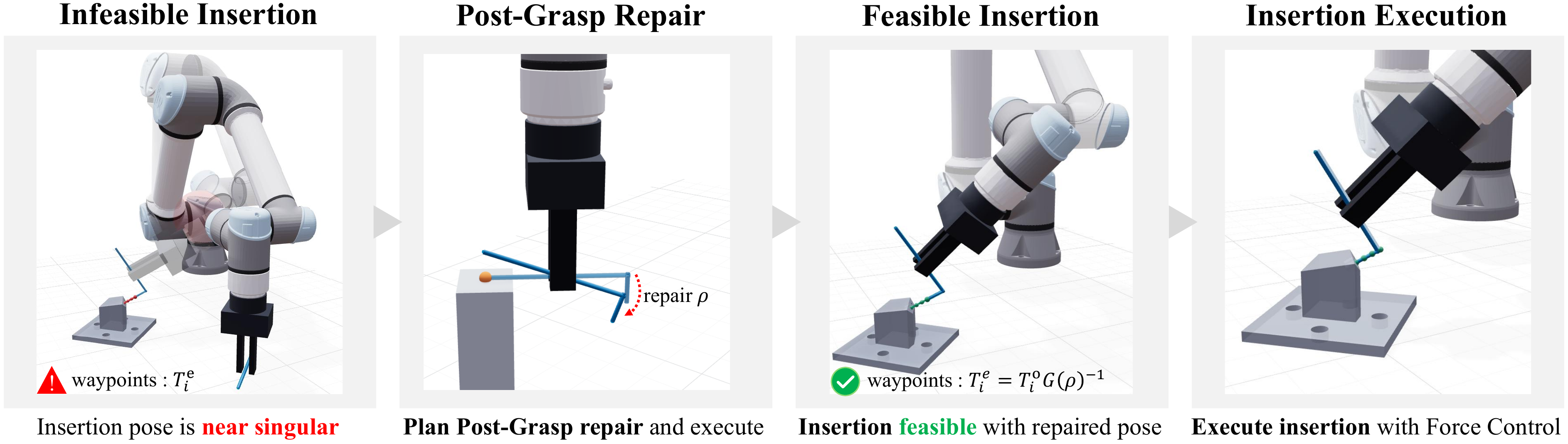}
        \captionof{figure}{\small{Overview of post-grasp kinematic feasibility repair. The nominal object-in-gripper transform leads to near-singular configurations along the prescribed insertion waypoints. The method selects a post-grasp reorientation $\rho$, realizes it through contact with the environment, and executes the resulting insertion under force control.
        }}
        \label{fig:overview}
    \end{minipage}
}

\begin{document}

\maketitle
\thispagestyle{empty}
\pagestyle{empty}
\begin{abstract}
A stable grasp does not guarantee kinematically feasible robotic insertion because the object-in-gripper transform may force the robot towards singularities or joint limits along the prescribed insertion path. We study post-grasp kinematic feasibility repair through object-in-gripper reorientation. Given an achieved grasp and a fixed insertion path, we seek a small reorientation that restores kinematic feasibility. Sequential IK can miss such candidates by following an unfavorable joint-space path, while the nonsmooth feasibility landscape makes the search computationally expensive. We evaluate candidates using a branch-aware IK graph that maximizes the minimum feasibility margin over the discretized insertion path and use a learned task-conditioned prior to improve query ordering. The selected reorientation is executed through tactile-based extrinsic manipulation. In UR5e simulations, the planner without learned ranking reduces mean reorientation over successful trials by 51.5\% compared with grid-based sequential IK. Adding learned ranking reduces this planner's mean planning time by an additional 51.1\%. Real-robot experiments validate the complete pipeline.  
\end{abstract} 
\section{Introduction}\label{sec:introduction}
Robots can perform repetitive pick-and-place tasks with high precision in fixed environments. However, achieving the same task becomes more difficult when the environment or setup changes. Depending on the available robot dexterity, task feasibility can vary with the initial grasp pose, robot configuration, and task requirements. Many tasks begin by grasping an object, and the resulting object-in-gripper transform constrains the subsequent robot motion. Thus, grasp stability is necessary but not sufficient for downstream task feasibility \cite{borst2004grasp, paolini2014data, lin2015grasp}.

Task-oriented grasp planning has been studied as a way to select grasps that improve downstream task feasibility \cite{vahrenkamp2010integrated, dang2012semantic, fang2020learning}.
However, being able to grasp an object does not guarantee downstream task feasibility, and a task-feasible grasp may not exist among the grasp poses reachable from the initial robot configuration.
Once the object is grasped, the relative transform between the gripper and the object is effectively fixed, and the downstream task must be executed only through the motion of the robot. This can make the task infeasible. 

To overcome the task infeasibility, extrinsic manipulation has been studied as a way to change the pose of a grasped object using contacts with the environment \cite{dafle2014extrinsic, liang2024robust}. These methods enable object reorientation through in-hand or extrinsic interactions. Recent developments in tactile sensors \cite{yuan2017gelsight, donlon2018gelslim} further make it possible to estimate the object motion relative to the gripper during such contact-rich reorientation \cite{kim2023simultaneous, kim2025texterity}. However, these extrinsic-manipulation methods often assume that the desired object reorientation is already specified. This leaves a separate planning problem: \textit{which object-in-gripper reorientation should be chosen to make the downstream robot motion feasible?}

In this work, we study post-grasp kinematic feasibility repair through object-in-gripper reorientation. Rather than assuming that a desired reorientation is given, we select it based on downstream kinematic feasibility. Each candidate transform induces a different robot motion for the same fixed object path. We evaluate each candidate with a branch-aware IK verifier that considers multiple IK solutions at each waypoint to assess kinematic feasibility along the prescribed insertion path. Because the resulting feasibility landscape is nonsmooth and costly to evaluate, we use an adaptive reorientation search with a learned task-conditioned prior for query ordering. Final acceptance remains determined by the branch-aware verifier.

The main contributions of this work are:
\begin{itemize}
    \item A post-grasp kinematic repair formulation that minimizes object-in-gripper reorientation while preserving the desired insertion path.
    
    \item A branch-aware max–min verifier using multiple IK solutions per waypoint to identify feasible paths missed by sequential IK.
    
    \item An adaptive reorientation search with learned candidate ranking that reduces IK-based feasibility evaluations.
\end{itemize}
\section{Related Work}\label{sec:related_work}
\subsection{Task-oriented grasp planning}
Robotic grasping requires accurate perception and control to identify stable and feasible grasp poses. Grasp quality is commonly evaluated using predefined metrics, including force-closure, wrench-space quality, or robustness under sensing and actuation uncertainty~\cite{mahler2017dex, maranci2024enabling}. Recent methods have advanced general grasp synthesis by learning grasp quality functions or generative grasp distributions from large-scale synthetic and real datasets~\cite{fang2023anygrasp, murali2025graspgen}. Task-oriented grasp planning extends this formulation by incorporating downstream task constraints into the selection of the initial grasp, for example through simulated self-supervision for tool manipulation~\cite{fang2020learning}. Recent vision-language and vision-language-action models further enable task-conditioned grasp selection from semantic task descriptions~\cite{black2025pi05, deng2025graspvla}. However, these methods primarily focus on selecting task-compatible initial grasps, rather than recovering task feasibility through post-grasp adjustment.
\subsection{In-hand manipulation and extrinsic dexterity}
Robots with simple grippers can have dexterity using external forces, such as gravity or forces generated by contact \cite{dafle2014extrinsic}. Extrinsic dexterity was used to adjust the pose of objects that are difficult to grasp due to occlusions \cite{zhou2023learning}. A recent study proposed primitive-based approaches to exploit environmental contacts to transform an initially ungraspable object into a graspable configuration before grasping \cite{yang2024learning}. Planar in-hand manipulation has been studied using motion cones, which characterize object motions admissible under frictional contact and support efficient sampling-based planning \cite{chavan2020planar} and nonlinear trajectory optimization \cite{perugini2025pushing}. Robust extrinsic-contact manipulation was proposed for adjusting the in-gripper pose of a grasped object while maintaining stable environmental contact under model uncertainty \cite{liang2024robust}. With the development of tactile sensing, recent studies have explored tactile estimation for in-hand and extrinsic manipulation, enabling robots to localize grasped objects or estimate contact states during manipulation \cite{lee2025trajectory, bauza2024simple, van2026simultaneous}. Other works have integrated tactile estimation with feedback control for extrinsic dexterity, allowing robots to track the pose of objects and regulate contact-rich reorientation using external contacts \cite{kim2023simultaneous, kim2025texterity}.

\subsection{Post-grasp task feasibility and IK verification}
Post-grasp manipulation is strongly affected by uncertainty in the object-in-hand state. Prior work has modeled this uncertainty from sensor observations and combined it with task-specific success models to predict whether placing, dropping, or insertion actions are likely to succeed~\cite{paolini2014data}. Related work in task and motion planning learns feasibility models to filter or guide high-level actions before invoking expensive geometric planners \cite{wells2019learning, ait2023learning}. These approaches treat the achieved object-in-gripper transform as a fixed or uncertain state when estimating task success. In contrast, we optimize a post-grasp adjustment to this transform to make a fixed downstream task path kinematically feasible.
\begin{figure}[t]
    \centering
    \includegraphics[width=0.625\linewidth]{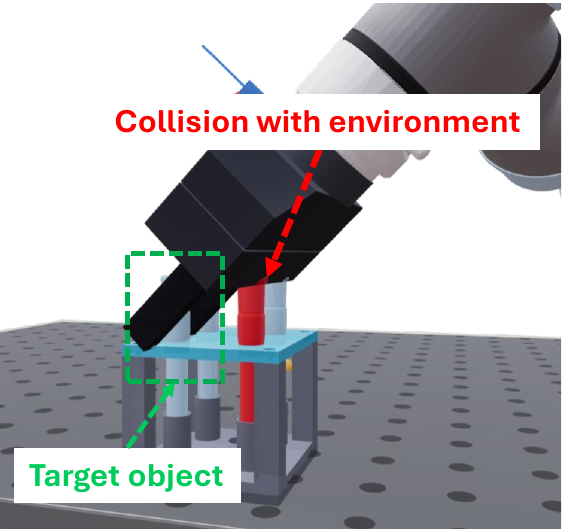}
    \caption{\small{Illustration of a constrained pickup scenario with four test tubes placed in a holder. The robot cannot reach an alternative grasp pose for the target object without colliding with a neighboring tube.}}
    \label{fig:motivation}
\end{figure}

Kinematic feasibility is commonly evaluated by solving IK along a given path and checking pose accuracy, joint limits, and joint-space continuity. However, tasks involving force control are particularly sensitive to configurations near kinematic singularities. Since redundant manipulators can admit multiple IK solutions for the same end-effector pose \cite{burdick1989inverse}, sequential IK can remain on an unfavorable solution sequence and falsely reject a feasible path. RelaxedIK~\cite{rakita2018relaxedik} generates motions from end-effector goals, whereas STAMPEDE~\cite{rakita2019stampede} and IKLink~\cite{wang2024iklink} select among multiple IK branches for a fixed end-effector path.
In contrast, we optimize the object-in-gripper reorientation while keeping the desired object path fixed, thereby changing the induced end-effector path. For each induced path, our branch-aware verifier then selects IK solutions from multiple IK branches.
\begin{figure*}[!t]
    \centerline{\includegraphics[width=0.95\linewidth]{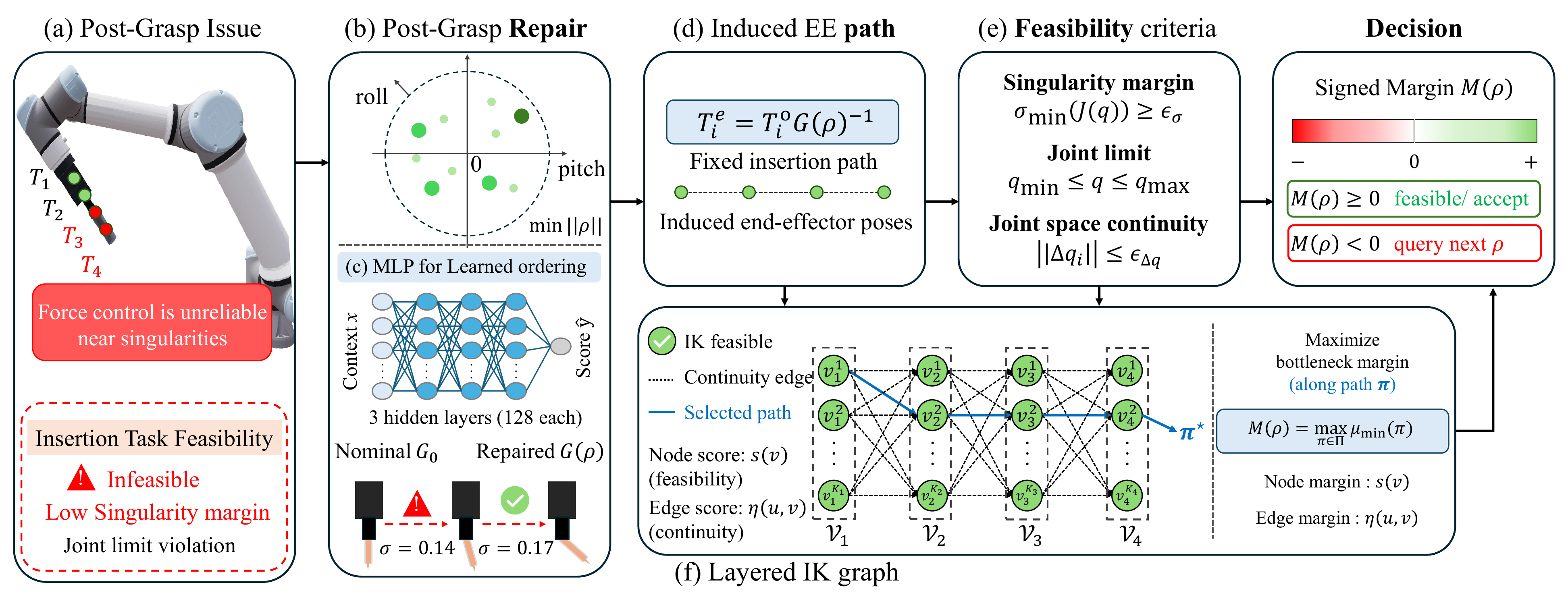}}
    \caption{\small{Overview of the proposed framework for post-grasp kinematic repair for robotic insertion via object-in-gripper reorientation.}}
    \vspace{-0.5cm}
    \label{fig:framework} 
\end{figure*}
\section{Problem Description}
As illustrated in \figref{fig:motivation}, environmental constraints may restrict the grasp poses available at pickup. For an $n_q$-degree-of-freedom robot with a grasp established under these constraints, the motion to follow a prescribed object insertion path may violate joint limits or approach kinematic singularities. The goal is to find a small object-in-gripper reorientation that makes the insertion motion kinematically feasible while keeping the desired object path fixed. The post-grasp state is characterized by the joint configuration $q_0 \in \mathbb{R}^{n_q}$ and the nominal object pose expressed in the end-effector frame, $G_0 \in SE(3)$. The prescribed insertion path consists of desired world-frame object poses $T_i^o\in SE(3)$, where $i$ indexes the $N$ waypoints. The joint configuration $q_s$ serves as the initial IK seed and the starting configuration for the joint motion approach. Candidate reorientations are evaluated with $q_s$ set to the initial post-grasp joint configuration $q_0$. After physical reorientation, $q_s$ is updated to the measured post-reorientation joint configuration $q_{\mathrm{post}}$.

An allowed adjustment of the object-in-gripper orientation is parameterized by $\rho\in\mathcal{P}$. The parameter space $\mathcal{P}$ depends on the manipulation mechanism, ranging from restricted in-hand rotations for a two-finger gripper to a richer reorientation space for a dexterous hand. In this paper, $\rho=[\rho_p,\rho_r]^\top$ represents local pitch--roll adjustments within $\mathcal{P}=\{\rho\in\mathbb{R}^2:\|\rho\|_2\le60^\circ\}$. Among feasible reorientations, we minimize $\lVert\rho\rVert_2$ to limit errors introduced during physical reorientation. The relative modification and the reoriented object-in-gripper transform are
\begin{equation}
  \Delta G(\rho)=
  \begin{bmatrix}
    R_y(\rho_p)R_x(\rho_r)&\mathbf{0}\\
    \mathbf{0}^\top&1
  \end{bmatrix},
  \qquad
  G(\rho)=G_0\Delta G(\rho).
  \label{eq:repaired_grasp}
\end{equation}
Here, $R_x$ and $R_y$ denote rotations about the nominal object-frame axes, with the object-frame origin assumed to remain fixed relative to the gripper. We assume that $\mathbf{0}\in\mathcal{P}$, so that $\rho=\mathbf{0}$ corresponds to the nominal grasp. The corresponding end-effector waypoints expressed in the world frame are
\begin{equation}
  T_i^e(\rho)=T_i^o G(\rho)^{-1},
  \qquad i=1,\ldots,N.
  \label{eq:end_effector_path}
\end{equation}

The robot first moves from $q_s$ to the selected first insertion configuration using the joint motion and then follows the Cartesian insertion waypoints in \eqref{eq:end_effector_path}. The joint motion approach is handled separately by the controller. Kinematic feasibility is considered only over the discretized insertion segment and requires pose accuracy, joint-limit satisfaction, a prescribed singularity margin, and continuity between adjacent waypoints.
\section{Branch-Aware Feasibility Verification}
\label{sec:repair}
For a given reorientation $\rho$, insertion feasibility is evaluated by searching for a sequence of joint configurations that follows the induced end-effector waypoints and satisfies the prescribed kinematic constraints.

\subsection{Kinematic Feasibility Score}
\label{sec:evaluator}
Insertion tasks are often executed using force or admittance control in Cartesian space. Such controllers can become unreliable near singular configurations. We therefore use the minimum singular value of the geometric Jacobian as the feasibility score. Pose accuracy and joint-limit are enforced as hard constraints. To avoid configurations near the joint limits, an IK candidate is treated as invalid if any joint lies within $3^\circ$ of either its lower or upper limit. For each valid IK candidate $q$ at waypoint $i$, we define the normalized waypoint feasibility score as
\begin{equation}
s_i(q,\rho)=\frac{\sigma_{\min}(J(q))-\epsilon_\sigma}{\epsilon_\sigma}.
\label{eq:waypoint_score}
\end{equation}
Here, $J(q)$ is the geometric Jacobian, $\sigma_{\min}(J(q))$ denotes its minimum singular value, and $\epsilon_\sigma$ is the prescribed singularity threshold. For an IK candidate satisfying the pose-accuracy and joint-limit constraints, $s_i(q,\rho)\ge0$ indicates that all waypoint-level requirements are satisfied.

\subsection{Sequential IK Verification}
\label{sec:seq_ik}
As a baseline, we first consider a sequential IK verifier that tracks the reoriented end-effector waypoints $\{T_i^e(\rho)\}_{i=1}^N$ by solving one IK problem per waypoint.
\begin{equation}
q_i^{\mathrm{seq}}=\mathrm{IK}\big(T_i^e(\rho);q_{i-1}^{\mathrm{seq}}\big), \qquad q_0^{\mathrm{seq}}=q_s.
\end{equation}
If IK succeeds at all waypoints, the sequential IK path is scored by its worst waypoint:
\begin{equation}
S_{\mathrm{seq}}(\rho)=\min_{i\in\{1,\ldots,N\}}s_i\big(q_i^{\mathrm{seq}},\rho\big).
\label{eq:seq_score}
\end{equation}
The reorientation is accepted only if $S_{\mathrm{seq}}(\rho)\ge0$ and adjacent insertion configurations differ by at most $\epsilon_{\Delta q}$, the maximum allowed joint-space displacement. The sequential verifier commits to the IK branch induced by the initial seed and can therefore reject a reorientation that is feasible through a different IK branch. This can also bias the search toward larger reorientations.

\subsection{Branch-Aware IK Verification}
To address the single-branch limitation of sequential IK, we propose a branch-aware verifier inspired by graph-based path-wise IK methods such as IKLink \cite{wang2024iklink}. For a fixed repair parameter $\rho$, we construct the layered IK graph shown in \figref{fig:framework}(f). Each layer $\mathcal{V}_i$ corresponds to the end-effector waypoint $T_i^e(\rho)$ defined in \eqref{eq:end_effector_path}.

For each waypoint, we solve IK for $T_i^e(\rho)$ from a deterministic set of initial joint configurations and retain solutions satisfying the pose-accuracy and joint-limit constraints. The initialization set includes $q_s$, the home configuration, the sequential IK solution, retained IK solutions from the previous waypoint, and random joint configurations drawn within the joint limits using a fixed random seed. The resulting finite candidate set and graph layer are
\begin{equation}
\mathcal{Q}_i(\rho)=\left\{q_i^k\right\}_{k=1}^{K_i}, \qquad \mathcal{V}_i(\rho)=\left\{v_i^k\right\}_{k=1}^{K_i}.
\label{eq:ik_graph_layer}
\end{equation}
Here, $K_i$ is the number of valid IK candidates at waypoint $i$. Each node $v_i^k$ represents $q_i^k$ and has score $s(v_i^k)=s_i(q_i^k,\rho)$.

For nodes in adjacent layers, we define the continuity margin as
\begin{equation}
\eta(v_i^k,v_{i+1}^{h}):=1-\frac{\lVert\operatorname{wrap}(q_{i+1}^{h}-q_i^k)\rVert}{\epsilon_{\Delta q}},
\label{eq:continuity_score}
\end{equation}
where $\operatorname{wrap}$ computes the shortest signed angular difference for each revolute joint. An edge is added only if $\eta(v_i^k,v_{i+1}^h)\geq0$. $k$ and $h$ index IK candidates in layers $i$ and $i+1$, respectively.

Branch-aware IK verification then becomes a path selection problem. A path $\pi$ selects one IK candidate from each waypoint layer, representing a complete IK branch over the insertion sequence. Let $\Pi$ denote the set of all such complete paths and $E(\pi)$ the set of edges in $\pi$. The minimum feasibility margin of a path is
\begin{equation}
\mu_{\min}(\pi)=\min\left\{\min_{v\in\pi}s(v),\min_{(u,v)\in E(\pi)}\eta(u,v)\right\}.
\end{equation}
A path is feasible if $\mu_{\min}(\pi)\geq0$, ensuring that every node satisfies the singularity threshold and every transition satisfies the joint-space continuity threshold.

We select the path that maximizes the minimum feasibility margin:
\begin{equation}
\pi^\star=\arg\max_{\pi\in\Pi}\mu_{\min}(\pi).
\end{equation}
Because edges only connect adjacent waypoint layers, this widest-path problem can be solved by dynamic programming \cite{lavalle2006planning}. Let $D_i(v)$ denote the largest minimum feasibility margin among partial paths ending at $v\in\mathcal{V}_i$. The first layer is initialized as
\begin{equation}
D_1(v)=s(v), \qquad v\in\mathcal{V}_1.
\end{equation}
For $i=2,\ldots,N$, the recurrence is
\begin{equation}
D_i(v)=\max_{u\in\mathcal{P}(v)}\min\left(D_{i-1}(u),\eta(u,v),s(v)\right),
\end{equation}
where $\mathcal{P}(v)$ is the set of predecessor nodes in $\mathcal{V}_{i-1}$ connected to $v$. The terminal node is selected as
\begin{equation}
v^\star=\arg\max_{v\in\mathcal{V}_N}D_N(v),
\end{equation}
and $\pi^\star$ is recovered by tracing the stored optimal predecessors. The branch-aware feasibility margin for $\rho$ is
\begin{equation}
M(\rho):=\mu_{\min}(\pi^\star)=D_N(v^\star).
\label{eq:branch_margin}
\end{equation}
The verifier returns $M(\rho)$ and the selected insertion path. If no path exists from the first layer to the final layer, $M(\rho)$ is set to $-\infty$.
\section{Verifier-Guided Reorientation Search}
Since each verifier call requires multi-solution
IK and branch-graph search, the key algorithmic question is how to choose repair candidates
efficiently.

\subsection{Adaptive Mesh Reorientation Search}
\label{sec:derivative-free_search}
Our goal is to find a small repair that passes branch-aware IK verification:
\begin{equation}
\begin{aligned}
\rho^\star & \in \operatorname*{arg\,min}_{\rho \in \mathcal{P}}
\quad \|\rho\|_2 \\
\text{s.t.} & \quad M(\rho) \geq 0.
\end{aligned}
\label{eq:repair_optimization}
\end{equation}
The branch-aware feasibility margin $M(\rho)$ is obtained through expensive verifier calls and may
be nonsmooth due to changes in the selected IK branch. We therefore use a derivative-free direct
search. Although our method is inspired by the search, poll, and adaptive-resolution structure of mesh adaptive direct search (MADS) \cite{audet2006mesh}, it uses task-specific update rules and does not implement the full MADS
algorithm. Across all adaptive-mesh experiments, we use mesh size $\Delta$ with initial value $\Delta_0=15^\circ$ and minimum value $\Delta_{\min}=0.25^\circ$. The reorientation magnitude is bounded by $\rho_{\max}=60^\circ$, and the search budget is $N_{\max}=96$ verifier evaluations. The mesh contraction and expansion factors are $\gamma_{\mathrm{dec}}=0.5$ and $\gamma_{\mathrm{inc}}=1.25$, respectively. At iteration $k$, $\mathcal{U}_k$ contains the positive and negative directions of an orthonormal basis rotated according to a fixed deterministic schedule.

Algorithm~\ref{alg:mads_repair} maintains a history $\mathcal{H}$ of evaluated pairs $(\rho,M(\rho))$. The search center is the evaluated candidate with the highest feasibility margin until a feasible repair is found, and the feasible candidate with the smallest reorientation magnitude thereafter. Radial refinement uses bisection between the origin and a feasible candidate, with a $0.25^\circ$ tolerance and at most 16 iterations. Feasible improvement denotes finding the first feasible repair or reducing the best feasible repair norm by more than $0.25^\circ$.

\begin{algorithm}[t]
  \caption{Adaptive mesh reorientation search}
  \label{alg:mads_repair}
  \begin{algorithmic}[1]
  \Require verifier $V$, domain $\mathcal{P}$
  \Ensure best verified repair or failure
  \State $\Delta\gets\Delta_0$, $k\gets0$, $\mathcal{H}\gets\emptyset$
  \State Evaluate $0$; \textbf{return} $0$ if feasible
  \State Evaluate candidates in $\Delta\mathcal{U}_0$; refine the best feasible candidate, if any
  \While{$|\mathcal{H}|<N_{\max}$ and $\Delta\ge\Delta_{\min}$}
      \State $k\gets k+1$; select center $z$
      \State Evaluate radial-search candidates; refine feasible candidates
      \State Evaluate candidates in $z+\Delta\mathcal{U}_k$; refine the best feasible candidate, if any
      \If{feasible improvement}
          \State $\Delta\gets\max(\Delta_{\min},\gamma_{\mathrm{dec}}\Delta)$
      \ElsIf{no feasible repair exists and the best infeasible margin improved}
          \State $\Delta\gets\min(\rho_{\max},\gamma_{\mathrm{inc}}\Delta)$
      \ElsIf{no record improved}
        \State $\Delta\gets\gamma_{\mathrm{dec}}\Delta$
      \EndIf
      \State \textbf{break} if no new search/poll candidate is evaluated and refinement yields no improvement
  \EndWhile
  \State \Return minimum-norm feasible repair in $\mathcal{H}$, or failure
  \end{algorithmic}
\end{algorithm}

\subsection{Learned Candidate Ranking} \label{sec:learned_candidate_ordering} 
Adaptive mesh reorientation search reduces verifier evaluations relative to dense-grid search, but its efficiency still depends on which candidates are queried first. A task-conditioned model is trained to rank candidate reorientations during the search step, and the branch-aware verifier determines their feasibility. The task context $x\in\mathbb{R}^{29}$ comprises $q_0$ (6-D), $G_0$ and the target hole pose (translation vector and first two rotation matrix columns), and five path parameters: approach distance, start/end insertion depths, and approach/insertion waypoint counts. The network assigns each candidate $\rho$ a score $\hat y_\theta(x,\rho)$ used to prioritize candidates for verifier evaluation.

The model takes the concatenation of the task context $x$ and candidate repair $\rho$ as input. The input features are standardized using the training-set mean and standard deviation. As shown in \figref{fig:framework}(c), the model has three 128-unit hidden layers with SiLU activations and outputs a scalar ranking score. The learning dataset was generated offline by running the adaptive mesh search described in Section~\ref{sec:derivative-free_search} without the learned prior. We ran the search on 600 unique task contexts and recorded all candidates evaluated by the verifier, yielding 33,429 labeled queries. The contexts were split into 420 training, 90 validation, and 90 test contexts. 

Using the verifier result and the repair magnitude, we assign each evaluated candidate a soft label $y(x,\rho)\in[0,1]$ that reflects its priority for verifier evaluation:
\begin{equation}
  y(x,\rho)=
  \mathbb{I}[M(x,\rho)\geq 0]
  \exp\left(-\frac{\|\rho\|}{\tau_\rho}\right)
  \exp\left(-\frac{R(x,\rho)}{\tau_R}\right).
  \label{eq:priority_score}
\end{equation}
$R(x,\rho)$ measures how much the repair norm exceeds the smallest feasible norm found among the evaluated candidates for $x$, allowing a $0.25^\circ$ tolerance:
\begin{equation}
    R(x,\rho)=\max\{0,\|\rho\|_2 -\min_{\rho':M(x,\rho')\geq0}\|\rho'\|_2- 0.25^\circ\},
\end{equation}
where $\rho'$ ranges over the feasible candidates evaluated for the same context $x$. If no feasible repair is found for $x$, all labels for that context are set to zero. The label favors small feasible repairs and applies an additional penalty to excess repair relative to the smallest feasible repair found for the same context. The scales $\tau_\rho=20^\circ$ and $\tau_R=10^\circ$ control these two penalties, respectively, with smaller scales producing stronger penalties. Using these labeled samples, the sample-weighted pointwise loss is

  \begin{equation}
    \mathcal{L}
    =
    -\frac{1}{\sum_{i\in\mathcal{B}} w_i}
    \sum_{i\in\mathcal{B}} w_i
    \left[
    y_i\log \hat{y}_{\theta,i}
    +
    (1-y_i)\log(1-\hat{y}_{\theta,i})
    \right],
    \label{eq:learned_ordering_loss}
  \end{equation}
where $w_i=2$ for feasible candidates and $w_i=1$ otherwise. We use Adam with learning rate $10^{-3}$, a batch size of 128, 120 epochs, and random seed 0. At each iteration, the network ranks pitch-roll repair candidates sampled in polar coordinates using 128 angular directions and up to 24 radii on the current mesh. The top-$4$ candidates are then evaluated by the verifier, while the poll and mesh-update steps remain unchanged.
\begin{table*}[t]
    \centering
    \caption{\small{UR5e insertion repair benchmark on 250 category-controlled test instances. Repair statistics are computed over successful trials for each method.}}
    \label{tab:simulation_main}
    \small
    \setlength{\tabcolsep}{4pt}
    \begin{tabular}{llccrrrrrr}
    \toprule
    ID & Method & Search & Verifier
    & Success (\%) & Repair (deg) & Median (deg)
    & Cand. evals & IK calls & Time (s) \\
    \midrule
    M0 & No repair
    & -- & Seq. IK
    & 20.0 & 0.00 & 0.00 & 1.00 & 6.0 & 0.03 \\
    
    M0-B & No repair
    & -- & Branch graph
    & 28.8 & 0.00 & 0.00 & 1.00 & 87.5 & 0.37 \\
    \midrule
    M1 & Grid-seq.
    & 5$^\circ$ grid & Seq. IK
    & 94.4 & 15.92 & 8.54 & 441.00 & 2646.0 & 16.37 \\
    
    M2 & Active-seq.
    & Ray active & Seq. IK
    & 94.8 & 11.98 & 6.88 & 59.47 & 356.8 & 2.03 \\
    
    M3 & Mesh-seq.
    & Adaptive mesh & Seq. IK
    & 94.8 & 11.68 & 6.56 & 73.53 & 441.2 & 2.57 \\
    \midrule
    M4 & Grid-branch
    & 5$^\circ$ grid & Branch graph
    & 94.4 & 13.00 & 7.07 & 441.00 & 39288.7 & 204.36 \\
    
    M5 & Active-branch
    & Ray active & Branch graph
    & 94.8 & 8.62 & 3.75 & 53.86 & 2034.0 & 8.69 \\
    
    M6 & Mesh-branch
    & Adaptive mesh & Branch graph
    & 94.8 & 7.72 & 2.00 & 62.56 & 2342.4 & 10.23 \\
    
    M7 & Learned-branch
    & Learned mesh & Branch graph
    & 93.6 & 8.08 & 3.70 & 35.69 & 1339.0 & 5.00 \\
    \midrule
    M8 & Seq.-IK NLP
    & SLSQP & --
    & 60.8 & 7.12 & 1.96 & -- & -- & 7.40 \\
    
    M9 & Branch-informed NLP
    & SLSQP & --
    & 89.2 & 10.30 & 4.38 & -- & -- & 10.87 \\
    \bottomrule
    \end{tabular}
    \vspace{-0.3cm}
\end{table*} 
\section{Experiments}\label{sec:Experiment}
The proposed repair problem is evaluated on the kinematic UR5e insertion task shown in \figref{fig:framework}(a). The robot must insert a grasped object into a target hole by aligning the object with the hole axis. The grasped object is modeled as a cylinder in simulation. We assume that the corresponding object-in-gripper transform 
has already been achieved before evaluating the insertion motion for each selected $\rho$. This assumption isolates the question studied in this work: whether object-in-gripper reorientation can restore kinematic feasibility of the downstream insertion path. All simulation experiments were run on an Intel Core i7-1360P CPU using PyRoki~\cite{kim2025pyroki}, with $\epsilon_\sigma$ and $\epsilon_{\Delta q}$ set to $0.14$ and $120^\circ$.
\subsection{Simulation Benchmark}
Table~\ref{tab:simulation_main} compares nine search--verifier combinations and two NLP baselines on the 250 insertion cases with UR5e. To avoid a test set dominated by easy cases from random sampling, we generated a larger pool of task cases and selected 250 test cases. Each test case consists of an initial robot configuration, a nominal object-in-gripper transform, and a target hole pose. The resulting test set covers five categories: already-feasible, repairable, near-singular, branch-sensitive, and unrepairable cases. The 250 benchmark contexts are disjoint from the 600 learning contexts and contain 50, 62, 62, 63, and 13 cases in the five categories above, respectively.

Success is the percentage of test cases with a feasible insertion trajectory. Repair is the norm of the selected pitch--roll repair $\rho$ in degrees. Cand. evals counts the evaluated repair candidates, IK calls counts IK solver calls, and Time is the wall-clock planning time. These three metrics are averaged over all test cases, whereas Repair is averaged over successful trials. M0 evaluates the nominal grasp without repair. M1--M3 combine sequential IK with grid, active ray-based, and adaptive mesh search, respectively. They test whether search alone can succeed with a single IK branch. M4--M6 use the same searches with a branch-graph verifier that considers multiple IK branches along the insertion path. M7 adds learned candidate ranking to M6 while retaining its adaptive mesh search and branch-graph verifier.

\subsection{Nonlinear Programming Baselines}
The feasibility-repair problem can also be formulated as a nonlinear program (NLP) that jointly optimizes a repair $\rho$ and insertion configurations $q_{1:N}$. By imposing kinematic constraints directly on repair and joint configuration, this formulation avoids differentiating the feasibility margin $M(\rho)$.
\begin{equation}
\renewcommand{\arraystretch}{1.12}
    \begin{array}{@{}c@{\quad}l@{\;}l@{}}
    \underset{\rho,\,q_{1:N}}{\operatorname{minimize}}
    & \displaystyle\|\rho\|_2^2 & \\[4pt]
    \text{subject to}
    & {}^{w}T_e(q_i)=T_i^e(\rho), & i=1,\ldots,N, \\
    & q_i\in\mathcal{Q}_{\mathrm{lim}}, & i=1,\ldots,N, \\
    & s_i(q_i,\rho)\ge0, & i=1,\ldots,N, \\
    & \|\operatorname{wrap}(q_{i+1}-q_i)\|_2\le\epsilon_{\Delta q},
    & i=1,\ldots,N-1, \\
    & \|\rho\|_2\le\rho_{\max}. &
    \end{array}
    \label{eq:nlp_baseline}
\end{equation}
Here, ${}^{w}T_e(q)$ denotes the end-effector pose in the world frame at configuration $q$, and $\mathcal{Q}_{\mathrm{lim}}$ denotes the joint-limit.  
 
The NLP remains nonconvex and its outcome can depend on the initial joint trajectory. Therefore, two NLP baselines differ only in how they initialize the joint trajectory. Sequential-IK NLP (M8) solves IK once per waypoint, using $q_0$ as the starting configuration at the first waypoint and the IK solution from the previous waypoint at later waypoints. Branch-informed NLP (M9) solves IK from multiple initial configurations, including the solution selected at the previous waypoint, $q_0$, home, and eight random joint configurations sampled once per repair start and reused across all waypoints. When constructing the initial trajectory for the NLP, M9 selects the IK solution at each waypoint that requires the smallest joint-angle change from the previously selected configuration.

Both use the initial repair candidates of the adaptive mesh search as NLP starting points, with at most 40 SLSQP iterations per start. Pose, continuity, and repair-domain derivatives are analytic; singularity derivatives use local finite differences with a step size of $10^{-4}$. Among candidates satisfying the stated constraints, the one with the smallest repair norm is selected.

\subsection{Repair Performance}
Without reorientation, branch-aware verification (M0-B) increases the success rate from 20.0\% to 28.8\% compared with sequential IK (M0), corresponding to 22 more successful cases. For each search strategy, branch-aware methods M4--M6 match the success rates of sequential IK methods M1--M3 (94.4\% for grid search and 94.8\% for the other searches) while requiring smaller mean repairs. Compared with grid search M4, adaptive mesh search M6 reduces IK calls by 94.0\% and increases success from 94.4\% to 94.8\%. 

The grid search methods (M1 and M4) missed one additional case with the $5^\circ$ grid but solved it when rerun with a $1^\circ$ grid. We did not rerun all 250 cases at $1^\circ$ because this would require approximately 25 times as many candidate evaluations and substantially increase computation time.

Relative to M6, M7 reduces candidate evaluations by 43.0\% and all-case planning time by 51.1\%, but success decreases from 94.8\% to 93.6\%, corresponding to three fewer successful cases. Over the respective successful trials, its mean repair is $0.36^\circ$ larger. Learned ranking thus reduces query cost with a slightly lower success rate and a higher mean repair.

Multi-IK initialization in M9 increases NLP success from 60.8\% (M8) to 89.2\%, with longer planning time and a higher mean repair over the respective successful trials. M9 remains below M6 (94.8\%) and M7 (93.6\%) in success rate. M8 and M9 mainly succeed on small-repair cases, lowering their reported averages; on jointly solved cases, M6 and M7 yield smaller repairs.
\begin{table}[t]
  \centering
  \caption{\small{Frozen-model sensitivity under controlled OOD variations.
  M6 and M7 achieved identical success in each condition. Calls denotes the mean number of verifier evaluations, and Red. denotes the percentage reduction achieved by M7 relative to M6.}}
  \label{tab:ood_sensitivity}
  \footnotesize
  \renewcommand{\arraystretch}{1.05}
  \begin{tabular*}{\columnwidth}{
    @{\extracolsep{\fill}}llcrrr@{}
  }
  \toprule
  OOD factor & Condition
  & \shortstack{Success\\(both)}
  & \shortstack{Calls\\M6}
  & \shortstack{Calls\\M7}
  & \shortstack{Red.\\(\%)} \\
  \midrule
  Goal position
  & Nominal
  & 20/20 & 69.15 & 34.60 & 50.0 \\

  & $+100$ mm
  & 18/20 & 67.50 & 41.10 & 39.1 \\

  & $+150$ mm
  & 17/20 & 67.50 & 49.55 & 26.6 \\
  \midrule
  Cylinder length
  & $-15\%$
  & 20/25 & 51.12 & 30.72 & 39.9 \\

  & $-30\%$
  & 20/25 & 39.44 & 25.00 & 36.6 \\

  & $-60\%$
  & 20/25 & 39.52 & 25.40 & 35.7 \\
  \midrule
  Object shape
  & Allen key (3)
  & 75/75 & 52.85 & 22.25 & 57.9 \\
  \bottomrule
  \end{tabular*}
\end{table}

\begin{figure*}[!t]
    \centerline{\includegraphics[width=0.925\linewidth]{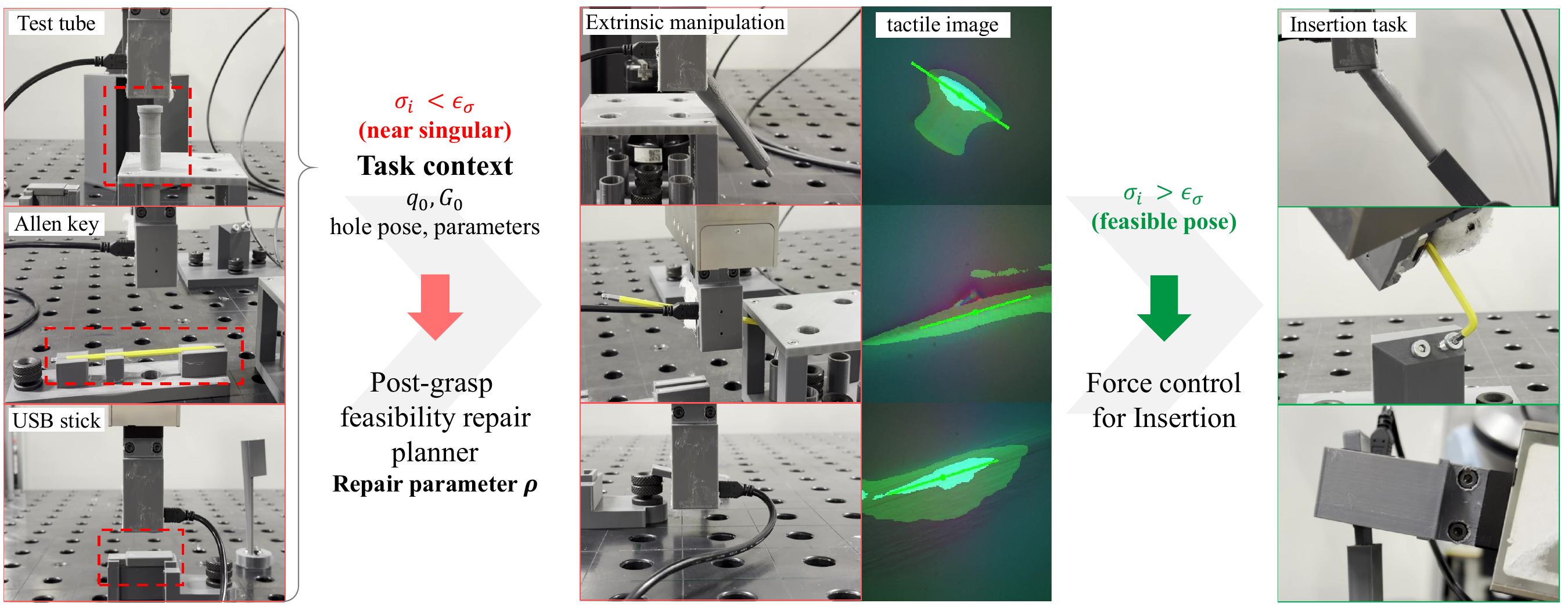}}
    \caption{\small{Real-robot experimental setup with three objects: a test tube, an Allen key, and a USB stick. The planned reorientation $\rho$ was executed by pushing the grasped object against the environment using tactile object-state feedback.}}
    \vspace{-0.3cm}
    \label{fig:exp_setup} 
\end{figure*}
\subsection{Out-of-Distribution Sensitivity}
Because M7 learns candidate ranking from a specific task distribution, generalization to arbitrary tasks is not claimed. We instead evaluate the frozen model without fine-tuning under controlled changes in insertion-goal position, cylinder length, and object shape, as summarized in Table~\ref{tab:ood_sensitivity}.

The reduction in verifier calls achieved by M7 diminished with increasing goal displacement but remained substantial across the tested length and shape changes. For Allen keys, this efficiency gain was accompanied by a small increase in mean repair angle, from $2.879^\circ$ to $2.998^\circ$. These results suggest that, within the tested conditions, the efficiency benefit of M7 is more sensitive to task-context shifts than to object-geometry changes.

\subsection{Real-Robot Experiment Setup}
The simulation evaluation assumes that the selected reorientation parameter can be achieved exactly before insertion. In practice, however, the commanded reorientation may not be achieved exactly. Therefore, the experiment evaluates the physical execution of a verifier-selected reorientation $\rho$ and the insertion feasibility of the measured post-reorientation state. This experiment evaluates the complete pipeline from $\rho$ selection to tactile-based reorientation, post-reorientation verification, and insertion. Repair targets were selected by task-specific versions of M7. The reorientation controller serves only as an execution mechanism and is not a contribution of this work. Existing methods, such as \cite{kim2023simultaneous, kim2025texterity}, could provide alternative mechanisms for achieving extrinsic dexterity.

As shown in \figref{fig:exp_setup}, we used a UR5e with a WSG-50 parallel-jaw gripper and a finger-mounted GelSight Mini sensor to manipulate a test tube, Allen key, and a USB stick. The robot pushed each grasped object against the predefined environment while a PD controller tracked a reference object-in-gripper trajectory from $G_0$ to $G^\star$ using tactile estimates of relative object motion.

Tactile-based reorientation may introduce rotational tracking error and translational slip and therefore may not achieve the commanded transform $G^\star$ exactly. After reorientation, we reran the branch-aware verifier using $q_{\mathrm{post}}$ as the initial IK seed and the estimated post-reorientation object-in-gripper transform $\hat{G}$. Insertion was attempted only when the verifier returned a nonnegative feasibility result; the robot then approached its first configuration using the joint motion.

For each object, we evaluated three different hole configurations, each repeated five times, resulting in 45 trials in total. Each configuration was selected such that the nominal insertion path approached a kinematic singularity. We report three outcomes: reorientation success, post-reorientation verification success, and end-to-end insertion success.
\begin{table}[t]
    \centering
    \caption{\small{Real-robot evaluation results for three objects.}}
    \label{tab:real_robot_results}

    \setlength{\tabcolsep}{2.8pt}
    \renewcommand{\arraystretch}{1.15}

    \resizebox{\columnwidth}{!}{%
    \begin{tabular}{lcccccc}
        \toprule
        & & &
        \multicolumn{3}{c}{Proposed pipeline (Successes/Attempts)} \\
        \cmidrule(lr){4-6}

        Object
        & Configs.
        & Trials
        & \shortstack{Reorientation\\success}
        & \shortstack{Post-action\\verifier pass}
        & \shortstack{End-to-end\\insertion} \\
        \midrule

        Allen key & 3 & 15 & 15/15 & 15/15 & 13/15 \\

        USB stick & 3 & 15 & 15/15 & 15/15 & 12/15 \\

        Test tube & 3 & 15 & 15/15 & 15/15 & 15/15 \\

        \midrule
        Overall & 9 & 45 & 45/45 & 45/45 & 40/45 \\

        \bottomrule
    \end{tabular}%
    }
\end{table}

\subsection{Real-Robot Results}
Table~\ref{tab:real_robot_results} summarizes the real-robot results. Reorientation success required an angular error below $0.5^\circ$ between the target and estimated object-in-gripper orientations. Insertion success required reaching the prescribed depth without grasp loss, controller abort, or violation of the safety force threshold. End-to-end success required successful reorientation, post-reorientation verification, and insertion. Nominal insertion without repair failed for all three objects. With repair, reorientation and post-reorientation verification succeeded in all 45 trials. End-to-end insertion succeeded in 40 of 45 trials, with the remaining five trials failing during insertion after successful reorientation and kinematic verification.

\section{Conclusion and Limitations}
We presented a post-grasp kinematic repair framework that searches for a small object-in-gripper reorientation while preserving the desired insertion path. The branch-aware verifier finds feasible IK paths missed by sequential IK, while adaptive search and learned ranking reduce expensive verifier calls. Real-robot experiments show that the selected reorientations can restore insertion feasibility through tactile-based extrinsic manipulation.The method requires a platform-specific singularity threshold and verifies feasibility only at discrete waypoints, without guarantees along the continuous insertion path. Future work will extend verification to continuous paths and investigate the five insertion failures, focusing on contact effects and reorientation uncertainty for closed-loop repair replanning.

\section*{Acknowledgement}
This research was supported by Pioneer Center for Accelerating P2X Materials Discovery (CAPeX), DNRF grant number P3 and Fabrikant Vilhelm Pedersen og Hustrus Legat.
\bibliographystyle{IEEEtran}
\bibliography{bibliography}
\end{document}